# Structural Material Property Tailoring Using Deep Neural Networks

Ryan Noraas[1] and Nagendra Somanath[2]
*Pratt and Whitney Engines Division, United Technologies Corporation, East Hartford, CT 06108, USA*

Michael Giering[3] and Olusegun Oshin[4]
*United Technologies Research Center, United Technologies Corporation, East Hartford, CT 06108, USA*

**Advances in robotics, artificial intelligence, and machine learning are ushering in a new age of automation, as machines match or outperform human performance. Machine intelligence can enable businesses to improve performance by reducing errors, improving sensitivity, quality and speed, and in some cases achieving outcomes that go beyond current resource capabilities. These advances are largely driven by the availability of large amounts of data and access to more powerful computational resources, enabling us to learn from ever more complex information towards engineering and automating current resource-intensive tasks and processes. Relevant applications include new product architecture design, rapid material characterization, and life-cycle management tied with a digital strategy that will enable efficient development of products from cradle to grave. Traditional methods of analysis have been based largely on the assumption that analysts can work with data within the confines of their own computing environment, but the growth of "big data", new compute paradigms and methodologies is changing that approach, especially in cases in which massive amounts of data are distributed across locations. In addition, there are also challenges to overcome that must be addressed through a major, sustained research effort that is based solidly on both inferential and computational principles applied to design tailoring of functionally optimized structures. Current applications of structural materials in the aerospace industry demand the highest quality control of material microstructure (e.g. especially for advanced rotational turbomachinery in aircraft engines) in order to have the best tailored material property. In this paper, deep convolutional neural networks were developed to accurately predict processing-structure-property relations from materials microstructures images, surpassing current best practices and modeling efforts. The models automatically learn critical features, without the need for manual specification and/or subjective and expensive image analysis. Further, in combination with generative deep learning models, a framework is proposed to enable rapid material design space exploration and property identification and optimization. The implementation must take account of real-time decision cycles and the trade-offs between speed and accuracy.**

## I. Nomenclature

| | | |
|---|---|---|
| $D$ | = | depth (channels) of input image |
| $H$ | = | height of input image in pixels |
| $K_t$ | = | stress intensity factor |
| $R$ | = | stress ratio, defined as minimum stress over the maximum stress |
| $\sigma$ | = | alternating stress, ksi |
| $W$ | = | width of input image in pixels |

---

[1] Materials Engineer, Materials and Processes Engineering.
[2] Principal Engineer, Engine Systems and Integration, Systems Optimization, Associate Fellow, AIAA.
[3] Technical Fellow, Research, Machine Intelligence.
[4] Research Scientist, Autonomous and Intelligent Systems Department.

## II. Introduction

Traditionally, the design methodology behind materials development has been very focused and isolated. It has been assumed that the toughest, lightest, strongest material would always be more beneficial to structural designers. Therefore, materials were developed independent of their intended use for optimal material performance rather than optimal structural performance or product application life cycle. This limited design philosophy often results in materials that over-perform in some macro level structural applications, resulting in higher costs, and under-perform in other situations, requiring the use of additional material and once again increasing cost. This inability of materials to meet proper multidisciplinary end product performance objectives ultimately results in massive inflations in product costs in terms of economics, energy demand, and raw material resources.

An important concept in materials science is structural-property-performance relationship. Developing materials that meet the required performance and property goes back to control processing conditions, structural and compositions of the materials. Traditionally, the American Society for Testing and Materials (ASTM) specified controlled experiments are conducted to isolate the effect of one variable at a time. However, variables often are correlated with each other. It is infeasible to isolate some variable for experimental testing. Data mining can help reveal hidden relations between large amount of materials parameters, processing conditions and their relations with dependent materials properties. Traditional ways of material development can be disrupted and reshaped by making the use of available data.

Li and Fischer [1] proposed the integrated materials and structural design (ISMD) paradigm shown in Fig 1. This methodology links material scientists and engineers working on the micro-structural scale with structural designers working on the macro-structural scale through the design values which are common in both fields. Material engineers develop new materials for specific mechanical performance such as compressive strength, compliance or tensile ductility, and structural designers use these constitutive composite material properties in the design of structural members. By facilitating cooperation at this level between these two communities, materials can be engineered and tailored to closely match the expected structural demands, thus increasing the efficiency of the overall material-structure system. This is highly relevant in the harsh environment of high speed turbomachinery subject to high operational temperatures and pressures and diverse mechanical loads.

Bhadeshia et al. [2,3] applied neural network technique to model creep property and phase structure in steel. Crystal structure prediction and local crystal planes is another area of study for machine learning-based material identification due to the large amount of structural data in crystallographic database using methods like K-nearest neighbor to identify localized material structure based on its nearest neighbors [4,5]. Machine learning is also applied for materials discovery by material constitutive composition, tailored constitutive design space for desired properties, which is essentially solving a constrained optimization problem. Rodemerck et al. [6] were able to find an effective multicomponent catalyst for low-temperature oxidation of low-concentration propane with a genetic algorithm and neural network. Morgan and Ceder [7] identified the interdisciplinary nature of machine learning and recognized the challenges for developing methods needed for material tailoring at the micro-level, including predicting physiochemical properties of constituents, modeling electrical and mechanical properties, developing more effective catalysts, and predicting crystal structure

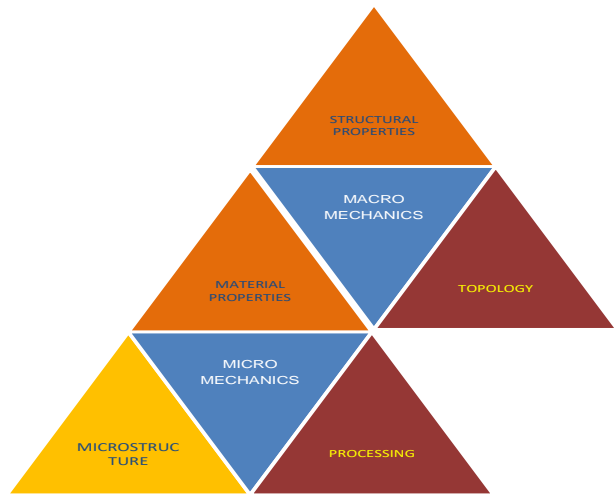

**Fig. 1  Integrated Materials and Structural Design (ISMD) Paradigm.**

Data in the material design space can be specifically driven by the scale as well as time. Furthermore, materials data tends to be complex and heterogeneous in terms of their sources and types, ranging from discrete numerical values to qualitative descriptions of materials behavior, test data and image data. The practical reality of missing data and uncertainties with the measured data is prevalent. With the emergence of "big data", methods to extract hidden information from complex image data and interpret resulting information for property estimation are becoming an increasingly important focus for materials design and development. Machine learning and statistically driven deep neural networks are used to learn from raw data collection, perform data preprocessing (filling in missing data, handling outliers, data transformation), feature engineering for feature identification, selection and extraction (principle component analysis), model selection, training, validations and testing. Artificial neural networks (ANNs) are inspired by the biological brain such that artificial neurons are connected to mimic the connection of neurons in the brain. Multiple hidden layers and neurons can add to the complexity of the neural network architecture. The strength of ANNs is that they are flexible and can represent any nonlinear and linear function. However, they need large amounts of training data and are prone to overfitting.

Materials design requires an understanding of how desired properties such as alloy yield strength, toughness, ultimate tensile strength, and fatigue life etc. are affected by intrinsic constitutive alloy microstructure, chemical composition, crystal structure, external processing, loading conditions and temperatures. Machine learning algorithms can derive the quantitative relations between the independent and dependent variables and hence make predictions with enough training data when physical models do not exist or are too complicated to derive. ANNs can also be used to predict constitutive relations. For instance, the constitutive flow behavior of an alloy is predicted with strain, log strain rate and temperature as input, and flow stress as output. Alloy ultimate tensile strength, yield strength, tensile elongation rate, strain hardening exponent and strength coefficient were also able to be predicted by ANN with a function of temperature and strain rate. Fatigue properties have always been among the most difficult to predict due to the high cost and time required for fatigue testing and the prevalence of structural failure caused by fatigue. Existing physical models are either lacking in generality or fail to give quantitative indications.

The inverse problem of identifying material properties based on constituents of the material is a more challenging multidisciplinary optimization problem because of the possibility of multiple solutions and the enormous dimensionality of the design space. Machine learning has shown promise in inverse materials discovery and design by rapid learning through deep neural networks and hence reducing the design space of relevance. Liu et al. [8] developed a machine learning method for the inverse design of Fe-Ge alloy microstructure with enhanced elastic, plastic and magnetostrictive properties. A systematic approach consisting of random data generation, feature selection, and classification was developed. This method was validated with five design problems, which involves identification of microstructures that satisfy both linear and nonlinear property constraints. This framework shows supremacy comparing with traditional optimization methods in reducing as much as 80% of running time and achieving optimality that would be difficult or would not be attained.

### III. Deep Generative Networks for Materials Design

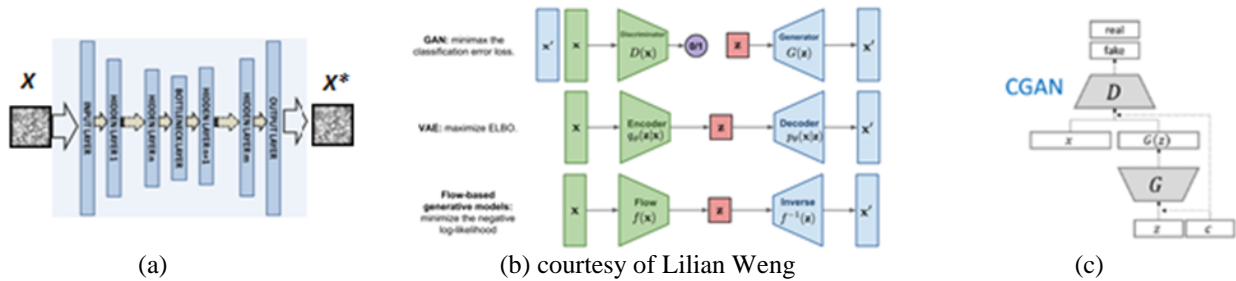

(a)　　　　　　　　　　(b) courtesy of Lilian Weng　　　　　　　　　　(c)

**Fig. 2** (a) A CNN with hidden layers, (b) A comparison of the GAN, VAE and Flow-based models, (c) A conditional GAN (cGAN)

In this paper, a family of deep convolutional neural networks (CNN), generative adversarial networks (GAN), variational autoencoder GAN (VAEGAN) and a Flow-based model will be introduced. Fig 2 shows the architectural schematics of these networks. Though they differ significantly in their architectural and functional response details, the commonality among these models is that they are all encoder-decoder assembly networks.

Convolutional neural network models (CNNs) [9-11] are comprised of an input layer, hidden layers and an output layer. The inter-layer connections, referred to as neurons have learnable weights and biases. Neurons receive scalar input pixel values. Convolution of the input layer with filter arrays of dimension less than the input space are followed by an optional non-linear transformation (often rectified linear units). The outputs of each step become the inputs to the subsequent layer. The model network and weights are differentiable with respect to the raw image input pixels and a specified loss function. CNN models commonly consist of three types of layers: convolutional, pooling, and fully-connected (dense). These layers are stacked in sequence to form the full model architecture. CNNs transform the input image, to the target response variable or class output.

These neural network models automatically handle the task of feature engineering and correlation of learned features with the specified response variable. Users do not need to specify up-front what is important in the images, thus eliminating human bias. Often times, materials engineers and scientists do not know with certainty which features are relevant to a given material property e.g., grain size or precipitate volume fractions. In these cases, it can be of exceptional value to be able to automatically determine critical features in an image or materials microstructure, without the need for manual specification of said features. As previously mentioned, using simplified parameters like average grain size may not numerically be adequate for prediction of complex material behavior. Higher order spatial interactions (N-point statistics), color, or texture descriptors can all be captured using CNNs [11-13] and used to predict material phenomena with improved accuracy over simplified material descriptors. This feature of CNNs, the ability to map image pixel values to a target response value, inevitably requires the use of training techniques to minimize risk for over-fitting, some of which are addressed in this paper.

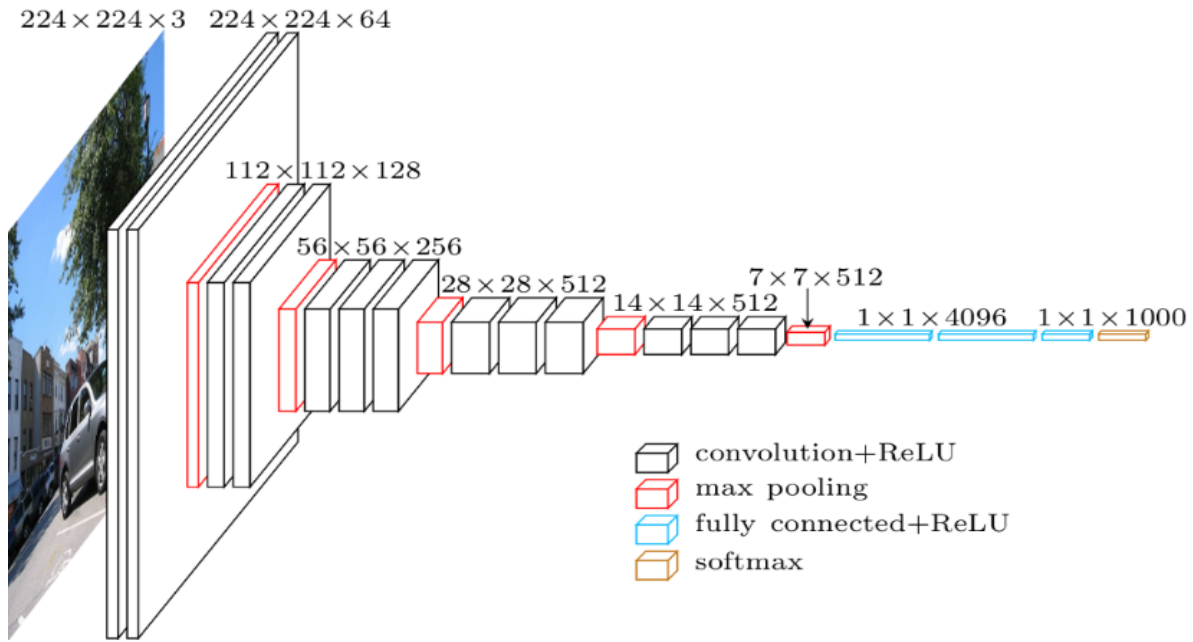

**Fig. 3 Schematic view of the VGG16 convolutional neural network**

The CNN is a powerful example of deep learning methods that use an encoder/decoder structure to train. The encoder maps images to an embedding space (a.k.a. latent space). Encoder/decoder based models can map from the embedding back to the image space. Fig. 4 shows a general schematic for deep networks leveraging an encoder/decoder architecture. For comparison, the schematic of the VGG16 in Fig. 3 only illustrates up to the embedding space. While many such models can learn an embedding, a required property for our work is that we can associate a single point in the embedding with each training input. This enables intelligent exploration in the embedding space and drives the choice of deep learning methods used.

It is a highly leveraged property that similar images cluster well and are separable in the learned latent or embedding space. One of the primary questions considered in this study is whether the subtle differences of microstructure across process and mixture variations are distinguishable in the embedding (or design) space. Once an image embedding has been trained, each point in the space can be decoded back to an image regardless of whether the point corresponds to a real input or not. Sampling the embedding between clusters of known materials and decoding enables the creation of synthetic microstructure images. The simulated microstructures are based on the non-linear relationships learned from the original input microstructure images. The design space exploration of the simulated microstructures are accurate and useful in reducing the expensive and time consuming process of manufacturing test materials.

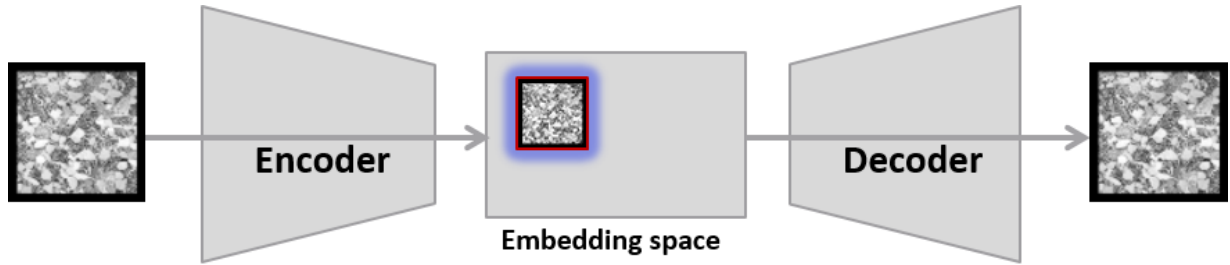

**Fig. 4 Schematic view of an autoencoder network**

Using a CNN model, it is possible to infer material pedigree robustly driven by thermomechanical processing during manufacture, predict material properties, assess quality control (pass/fail), and detect outlier microstructures in an automatic fashion using images. The value of these models can be further realized if linked into the design phase or with an optimization engine that can discern location-specific properties and model-based material definition. Adoption of deep generative models such as GANs may further enable materials design via generation of random, synthetic microstructure images for assessing effects of microstructural uncertainty on model predictions. It is envisaged that linking of various deep convolutional and generative neural network-based models (CNN [14], GANs [15,16], Autoencoders [17], Flow based models [18]) will enable data-driven materials design/exploration approaches using large historical databases built over time. If detailed material pedigree, properties, and microstructure images can be archived and made available to these machine learning models, benefits and sustained competitive advantage can be immediate. For example, a new or improved material can be rapidly optimized with respect to chemistry, processing, microstructure, and properties within this proposed framework. Synthetic microstructure images can be generated conditioned on design requirements which, in turn, would feed into the CNN image model to predict material properties (tensile, fatigue, creep, etc.). An optimization engine running underneath would iteratively adjust physical knobs in the manufacturing process (chemistry, forging or heat treat parameters) to identify the set of conditions that will theoretically generate an optimized material system possessing the user-specified requirements.

The objective of this study are: (1) demonstrate the capacity of convolutional neural networks for microstructure classification and property prediction in Ti-6Al-4V, an alloy used in the jet engine industry known for its high strength to weight ratio, low density, and exceptional corrosion resistance. (2) demonstrate the behavior of CNN models developed to perform classification of over 19 different heat treat pedigrees of the Ti-64 alloy, (3) demonstrate material design space exploration and the ability to predict high cycle fatigue, (4) Evaluate the use of deep generative models (conditional GANs, Glow), (5) Propose a deep learning based framework for linkage of various deep learning models and integration into an exploration/optimization design engine.

## IV. Results

**A. Microstructure Image Classification and Fatigue Prediction**

High cycle fatigue (HCF) is one of the primary causes of turbine-engine failure in aircraft applications [21]. Fatigue issues continue to lead to unpredictable failure in components due to crack propagation under complex loading conditions. Current research interest in structural aerospace alloys, mainly physics-based crystal plasticity models are being developed to understand local microstructure effect on small crack growth and crack initiation. Alloy microstructure can vary significantly during thermal and mechanical processing of the material required to achieve specific properties.

The main types of microstructures are (1) lamellar –formed after slow cooling from the heat treat solution temperature or in the mill-annealed state, (2) equiaxed –formed after deformation in the alpha-beta field, consisting of primary alpha (hexagonal close packed) dispersed in a beta matrix phase (body centered cubic high temperature phase). The dissolution of primary alpha phase is a thermodynamic and kinetic process controlled by the solution heat treat temperature. The lamellar microstructures are characterized by low tensile ductility, moderate fatigue, and good creep properties. The equiaxed microstructure provides a better balance of strength, ductility and fatigue properties [19-20].

Smooth [Smooth in this case refers to the physical fatigue specimen type, not the data. It can be smooth ($Kt=1$) or notched ($Kt > 1$)] high cycle fatigue data (75F, $Kt=1$, $R=0.4$) was available for 12 of the 19 Ti-64 pedigrees and was used in training a separate CNN model. As standard practice, individual parametric regression models were fit to each pedigree's fatigue data using only a single predicting variable ($1/\sigma$) and intercept parameter. This simple model fits most data sets with mean absolute error ranging from 0.2 – 0.4 in $\log_{10}$ cycles to failure; but does not explicitly take into account microstructure, which restricts model predictive capability. The target value of the CNN model was made to be the same as the fatigue regression model (i.e. predicting mean behavior, not specimen level).

Conventional standards for microstructure analysis have largely relied on domain expertise and visual conformity to established materials specifications. Often, these analyses are subjective, time-consuming and are not scalable to large, repetitive datasets. In this work, convolutional neural networks were used to classify microstructure images according to heat treat. The primary approach was to infer the micro constituent grain clusters that result from the heat treat that would enhance superior fatigue characteristics of the alloy system.

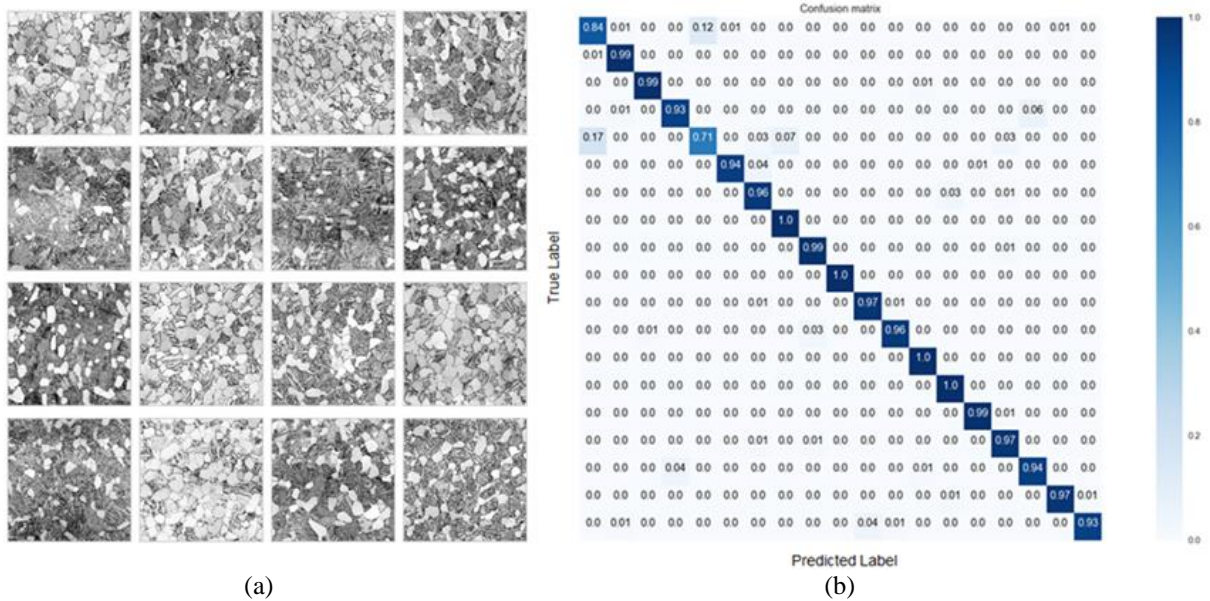

**Fig. 5 (a) Random sampling of Ti-6Al-4V images. (b) Confusion matrix showing classification results.**

Figure 5a shows a random sampling of images from the training dataset, containing 19 different material pedigrees, differing only by heat treatment. The confusion matrix (Fig. 5b) for the validation data set (30% withheld) is also shown, indicating >95% average accuracy on the reserve data set. During training of the CNN, random rotations, flips, and contrast augmentations were performed to reduce model sensitivity and risk of overfitting.

To predict fatigue values from image data, the VGG16 model architecture was modified (Fig. 6) and fine-tuned on a Ti-64 image dataset of 14,000 224x224 images by allowing parameter updates on the fully-connected layers. The architecture was developed to accept two inputs: a 224x224x3 image and a scalar value for the desired test stress (ksi).

Figure 7 compares target and predicted fatigue curves for 11 pedigrees of the Ti-6Al-4V using the above model architecture. The data was split such that 80% was used to train and validation was performed on the remaining 20%. It can be observed that the model is able to accurately predict fatigue for all tested pedigrees. The results show the model is able to generalize to new samples and captures the overall trend well. From a mechanistic standpoint, the model is forced to learn how to interpret the dual phase titanium microstructures. Empirically, there is a strong relationship between primary alpha volume fraction and grain sizes with fatigue capability. Primary alpha volume fraction, grain sizes, and secondary alpha lathe widths are resultant from materials processing, with strong dependence on solution temperature and cooling rate. Review of some of the misclassifications, as noted above, revealed that the mistaken microstructure classes are virtually indistinguishable to the trained eye. It is also noteworthy to mention that although the class labels are known with certainty, there is no guarantee that the resultant microstructures are statistically different. In which case, the model should not be able to accurately distinguish them.

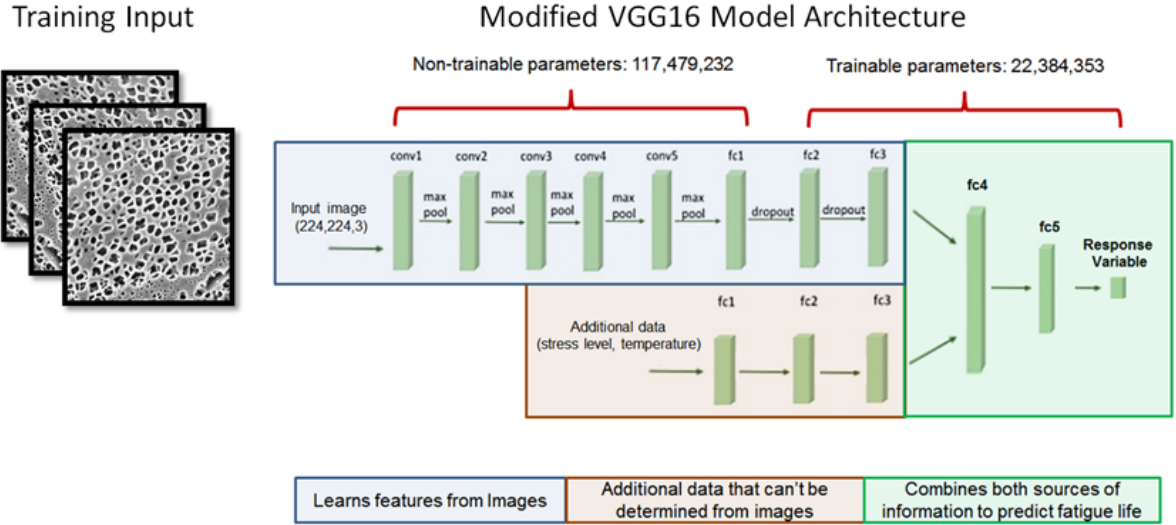

Fig. 6  Schematic view of modified VGG16 architecture, accepting two inputs (image and vector).

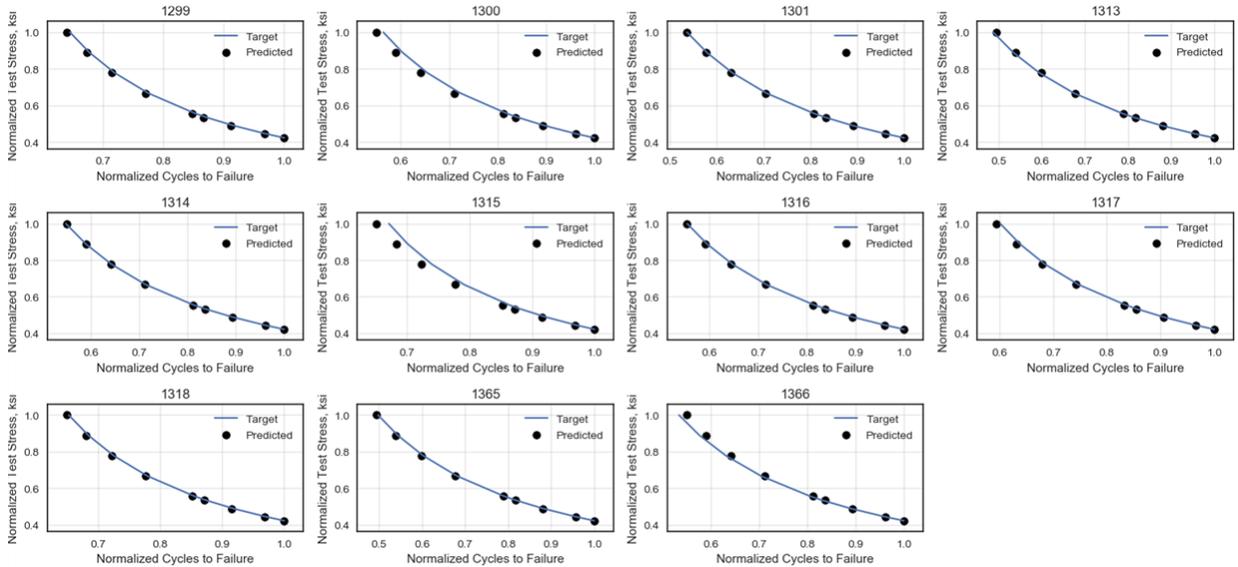

Fig. 7  HCF model predictions for 11 pedigrees of Ti-64A1-4V.

### B.  Material Design

It has been demonstrated above that CNN models can not only classify microstructure images, but also predict material property from constitutive microstructure images. We expect additional value to be available by leveraging unsupervised deep learning models (autoencoders), which can be conditioned, and learning transferrable features, especially grain boundaries and slip plane information (using image arithmetic [3]). To test our intuition, we apply Principal Component Analysis (PCA) to the output of the final fully connected layer of the VGG16 model, reducing the dimensions in order to visualize patterns in the data. Fig. 8 shows the resulting visualization. It is observed that the data can be split into three distinct regions corresponding to groupings of material pedigrees with High, Moderate and Poor fatigue strength and properties.

We also observe that fatigue strength decreases as the pedigrees move from left to right in the PCA space as indicated by the figure. It stands to reason, therefore, that the design of new materials should take into account the distribution of data in a representative latent space. Also, based on previous results, we expect that, provided enough well-labelled training data (chemistry, processing, microstructure, properties), it is possible to learn a low-dimensional representation of the material design space and enable rapid material system evaluation (exploration), as well as optimization with respect to certain specific material properties (like poisons ratio, moduli, creep behavior, etc.) that can be tested using ASTM test procedures. Specifically, with the generative capability of autoencoder models, synthetic microstructure images can be interpolated and generated based on the conditioning of the model.

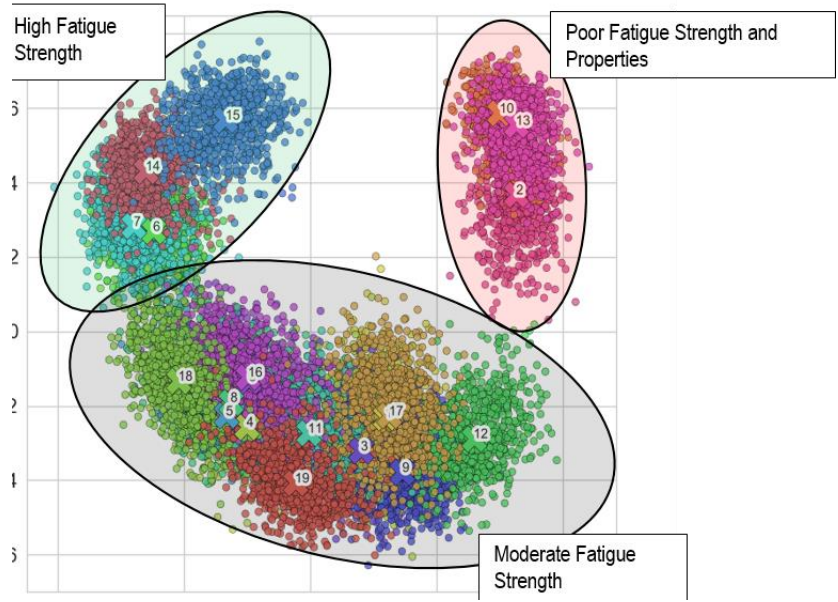

**Fig. 8 Principal Component Analysis (PCA) of microstructure image data from the final fully connected layer of the VGG16 network. We observe that the data is split into 3 regions corresponding to groupings of High, Moderate and Poor fatigue strength.**

Utilizing the generative capability of the variational autoencoder model (VAE), synthetic microstructure images can be produced with respect to labelled attributes (conditioning). For example, a microstructure image could be generated that has strong creep resistance, tensile strength, or a mixture of both. These synthetic images can be fed into validated CNN models for property or process prediction, and so an optimization loop can be set up to explore new design spaces. Variation and uncertainty in microstructures can be translated into uncertainty in model predictions by taking advantage of the stochastic sampling capability of variational autoencoders. A flow diagram of the process is shown in Fig. 9.

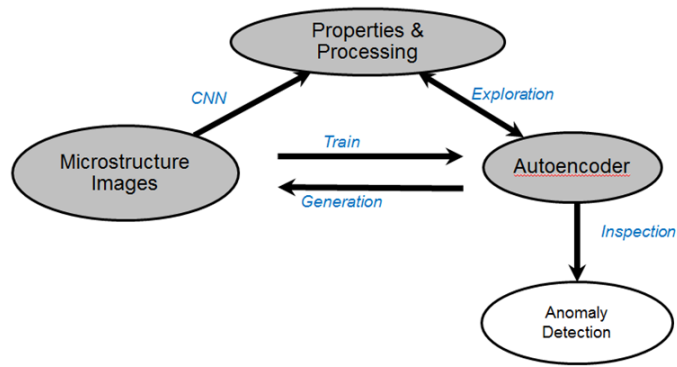

**Fig. 9 Flow diagram for property identification and training using a variational autoencoder model.**

*Glow and Conditional GAN (cGAN)*
One inherent limitation of the variational autoencoder model is that there is a degree of blurring in the output images. In [18], Kingma and Dhariwal introduced the Glow model, which was shown to generate clearer images and meet our embedding criteria. In particular, Glow provides exact latent-variable inference and provides a useful latent space for exploration. The Glow model uses a flow-based architecture, which scales well to large images and requires relatively fewer examples to train compared to other generative networks.

Once the Glow model is trained from existing data, generating new images involves sampling from random points in the latent space of the trained model. As expected, the result of this sampling would be a random image. However, it might be useful to direct the generation of pedigree images which exhibit certain known qualities. To generate images from a known pedigree, we project (encode) examples from that pedigree into the latent (embedding) space, and then limit sampling to the local latent region of the encoded examples. This pedigree-based image generation can also be achieved by interpolating in the latent space between any two examples of the pedigree of interest. Fig. 10 illustrates this sampling. The blue dots represent the interpolation path between the two images, which are at the ends of the path. These images are first projected into the latent space and are used to direct the sampling. (In the case of pedigree-based sampling, the projected images can belong to the same pedigree, but the interpolation principle is general). Sampling is then done along the path to generate new microstructure images from the represented pedigree. Fig. 11 shows examples of input micrograph images used to direct the image generation, and the generated micrograph images using this method.

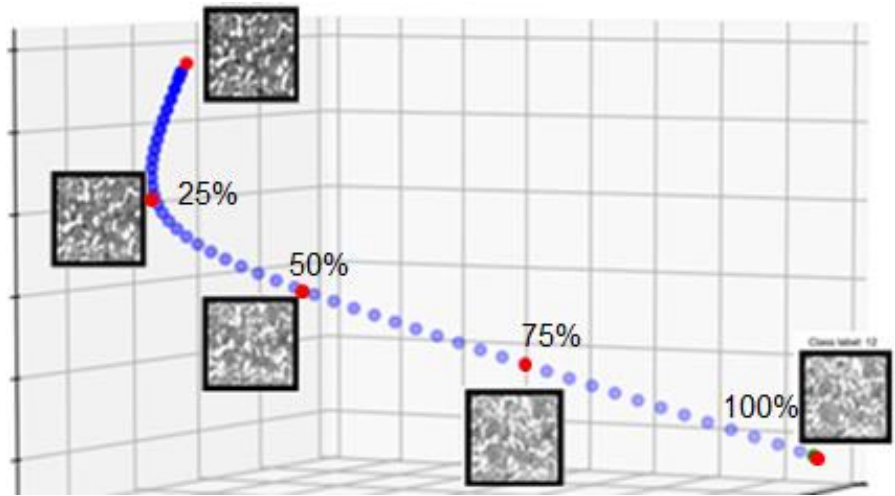

**Fig. 10 Illustration of image generation by interpolating between two images. This is useful for generating image within and between known regions in the latent space.**

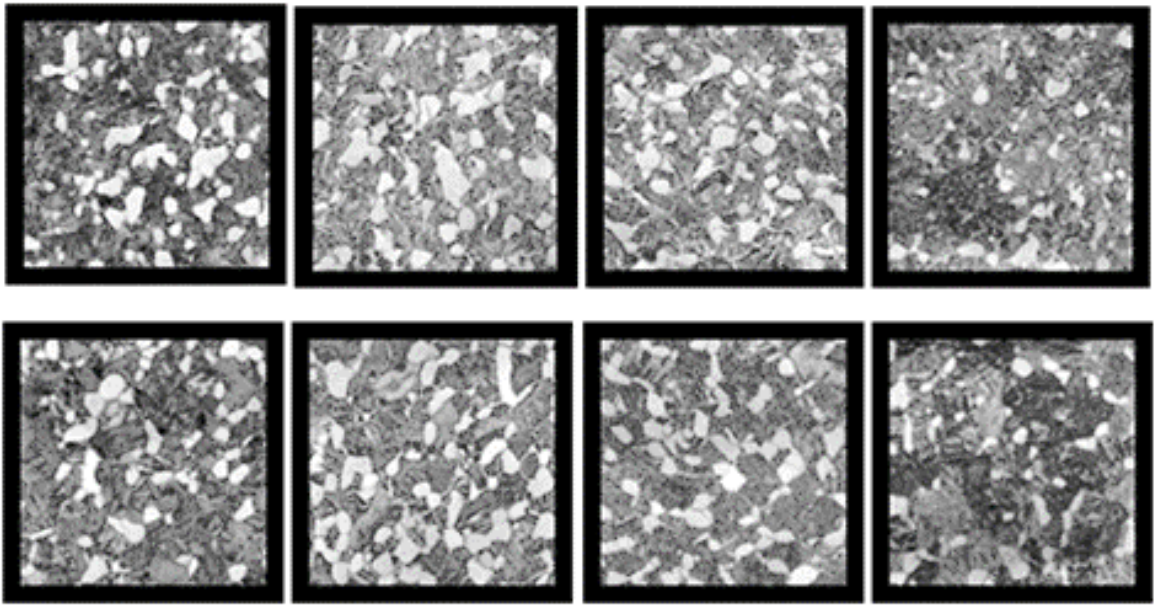

**Fig. 11** *Top row*: **Glow model generated micrographs of four Ti alloy pedigrees.** *Bottom Row*: **Input micrographs from the same pedigree classes, for reference.**

To generate high quality images, the Glow model requires a considerable amount of time to train. With insufficient training time, generated images exhibit subtle but still noticeable differences from input images. To accurately estimate the fatigue strengths from generated micrograph images, it is essential to improve the quality of the output image. To do this, we employ a conditional GAN (cGAN) [17], which learns a mapping of the output image to be improved to its equivalent high quality image. This results in a pipeline for microstructure image generation illustrated in Fig. 11.

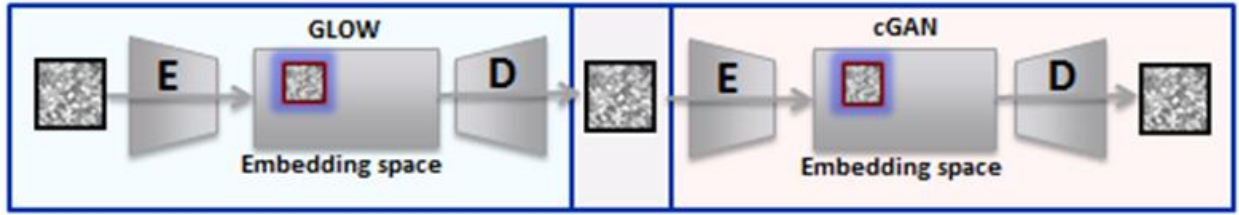

**Fig. 12 Model flow of data to learn an embedding space with the Glow model and improve the micrograph grain textures via a cGAN model. *E* represents the encoder, and *D*, the decoder.**

To evaluate the outputs of this approach, a range of objective and subjective metrics are used: (1) domain expert inspection, (2) classification of the model output with the VGG model described earlier, and (3) interpolation in the embedding space between pedigree clusters, for which the intermediate state is known. Domain experts evaluating the images from the Glow-cGAN process believe the structure characteristics and image quality to be within the normal range of what is observed in practice. In addition, classification of the model outputs were very similar to results on the raw input data, lending support that variation in pedigree structure and texture is preserved. Small artifacts and local blurring slightly decrease the performance of the automated image classification algorithm.

*Generating new materials*

To generate new materials, we need to explore the embedding space of the trained Glow model, navigating towards regions that represent desirable material design properties. However, it is necessary to validate this exploration. The PCA visualization in Fig. 8 provides an intuition for the groupings of the microstructure classes, and also suggests a direction for navigation towards desirable qualities. For our navigation experiments, we use the more sophisticated T-Distributed Stochastic Neighborhood Embedding (t-SNE) [22] to group the microstructure classes and determine directions for exploration. In order to validate this exploration, we generate pedigree samples from known regions in the latent space. These regions represent certain pedigree classes. Using the visualization in Fig. 8, two "starting" pedigree classes are selected along with an additional "leave-out" class, where the leave-out class lies in the path between the starting classes as indicated by the visualization. Validation involves training the model, without the leave-out class, then interpolating between the starting classes, generating synthetic microstructures in the space between them. Our expectation is that resulting interpolation would produce identical microstructures to the leave-out class.

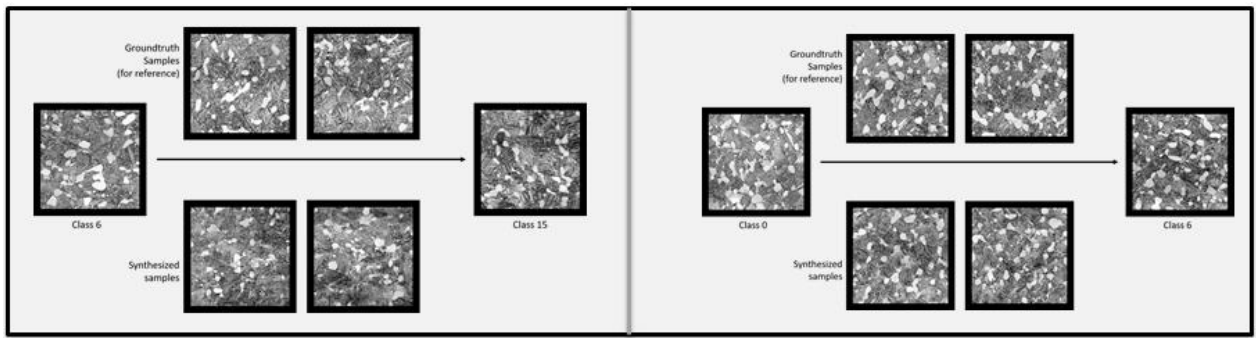

**Fig. 13 Results of interpolation and generation of images from unseen pedigrees. Groundtruth samples are above the arrows, while synthesized samples are below the arrows.**

Figure 12 compares the synthetic microstructure images with the ground truth for 2 different interpolations. Ground truth samples (above the arrows) are shown for reference and are representative of the intermediate class between the "starting" samples. Below the arrows are images synthesized from the left out class by interpolating between the starting samples. Qualitatively, it can be observed that the synthesized samples show similar features to the ground truth class that lies between the two starting classes. Generally, we expect that interpolation will be most accurate in regions of dense observational support.

For further validation, the generated images are run through a CNN trained to predict fatigue response. In Fig. 13, ground truth fatigue curves are provided for examples of the two starting pedigree classes, as well as samples generated from the interpolation of the starting class samples. It can be observed that the interpolated images produced a fatigue curve that is approximately between the two groundtruth curves, providing confidence for our intuition of the latent space, and further support that model and framework are able to interpolate between known pedigrees of material. The Glow model was successful in predicting within engineering judgement (2 – 3%) fatigue characteristics from observed engineering tests carried out.

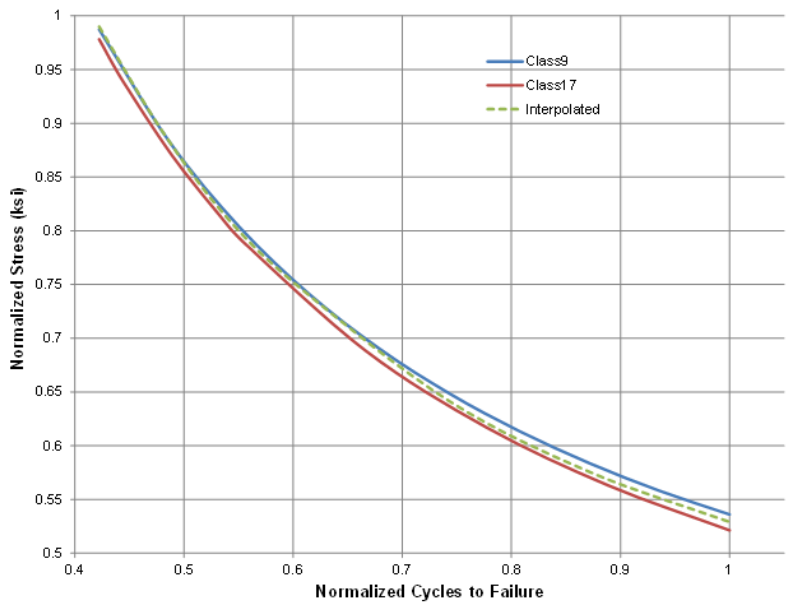

**Fig. 14 Predicted fatigue curve for image generated from interpolation.**

## V. Conclusion

A deep learning-based method for microstructure generation and material optimization was presented. The methodology is particularly useful for new material design and optimization, since currently the time between research and development of new materials and incorporation into industrial applications can span decades. The method is hinged on the availability of high quality data, including but not limited to material composition, quality, processing, microstructure, and performance. The present work exhibits a complete design methodology approach for developing sustainable infrastructure materials. At its core is the integrated structural and materials design paradigm. This design platform links together the varying scales involved with infrastructure materials design and implementation, ranging from microscale material tailorability to macroscale structural application. The specific material within this study represents a new class of materials for future aerospace material systems. The interdisciplinary integration of materials engineering, structural design, and computer science principles will enable rapid materials development, optimization, and manufacture while improving quality and performance.

In the future, we aim to further validate the synthesized microstructures and our approach to materials design optimization to predict fatigue of the interpolated microstructure from the synthesized images. This would reveal the extent to which the generative models are able to capture and synthesize nonlinear characteristic structural properties from training images. We intend to drive material design to user specifications of manufacturing processing route such as heat treat, chemistry, cooling rates, so the system can generate the resultant microstructure images.

## Acknowledgments

The authors thank Pratt and Whitney and United Technologies Research Center for providing a cohesive and collaborative environment, and sustained investment towards this work.